\documentclass[11pt]{article}

\usepackage[preprint]{acl}
\usepackage{times}
\usepackage{latexsym}
\usepackage{algorithm}
\usepackage{algpseudocode}
\usepackage{amsmath,amssymb}
\usepackage{booktabs,multirow,tabularx,colortbl}
\usepackage{tcolorbox}
\usepackage[T1]{fontenc}

\usepackage[utf8]{inputenc}

\usepackage{microtype}

\usepackage{inconsolata}

\usepackage{graphicx}

\title{MMKG-R1: Constructing Multimodal Knowledge Graphs via Reinforcement Learning}

\author{Zongyu Wu\thanks{Core contributors.}, Yilong Wang\footnotemark[2], Xiaochen Wang, Minhua Lin, Yuan Zhong,\\
{\bf Fenglong Ma, Xiang Zhang, Suhang Wang}\\
        The Pennsylvania State University\\
        \texttt{\{zongyuwu,yvw5769,xcwang,mfl5681,yfz5556,fenglong,xzz89,szw494\}@psu.edu}
        }

\begin{document}
\maketitle
\begin{abstract}
Multimodal knowledge graphs (MMKGs) organize information across different modalities to support multimodal Graph Retrieval-Augmented Generation (RAG), which aims to improve the performance of multimodal large language models (MLLMs). The effectiveness of these multimodal Graph RAG methods largely depends on the quality of the constructed MMKGs. However, most existing methods construct MMKGs without explicit guidance on graph quality, while reinforcement learning (RL)-based methods that rely on final-answer correctness may provide only indirect feedback on MMKG quality. To address these limitations, we propose MMKG-R1, a novel MMKG-quality-driven RL framework for training an MLLM as a multimodal knowledge constructor. To be specific, MMKG-R1 introduces multiple quality-aware rewards to directly evaluate both the quality of information extracted from individual knowledge sources and the quality of the assembled MMKG. These rewards are then used to guide the RL training. Experiments across multiple datasets show that MMKG-R1 constructs MMKGs that improve downstream question-answering accuracy.
\end{abstract}

\section{Introduction}
Multimodal large language models (MLLMs)~\citep{liu2023llava,Liu2024llava1.5,bai2025qwen3vl,mllmsurvey} have been widely adopted across various domains. Despite their effectiveness, they still suffer from outdated knowledge and hallucinations~\citep{bai2024multimodalllmhallucination,li2023pope}, especially when answering questions that require recent or unseen information. Retrieval-Augmented Generation (RAG)~\citep{Lewis2020RAG,gao2024llmrag,abootorabi2025mmragsurvey} is an established approach for reducing hallucinations in (M)LLMs~\citep{touvron2023llama1,wang2024qwen2vl} and enabling them to access and utilize external knowledge. As knowledge is often multimodal, including information like texts and images, recent works~\citep{guo2025raganything,dai2026mg2rag} start constructing multimodal knowledge graphs (MMKGs)~\citep{liang2024mmkgsurvey} to support multimodal Graph RAG such as MMGraphRAG~\citep{wan2025mmgraphrag}. By incorporating both semantic information in each modality and cross-modal associations, these methods aim to further improve retrieval quality and downstream question answering performance. 

\begin{figure}[t]
    \centering
    \includegraphics[width=0.9\linewidth]{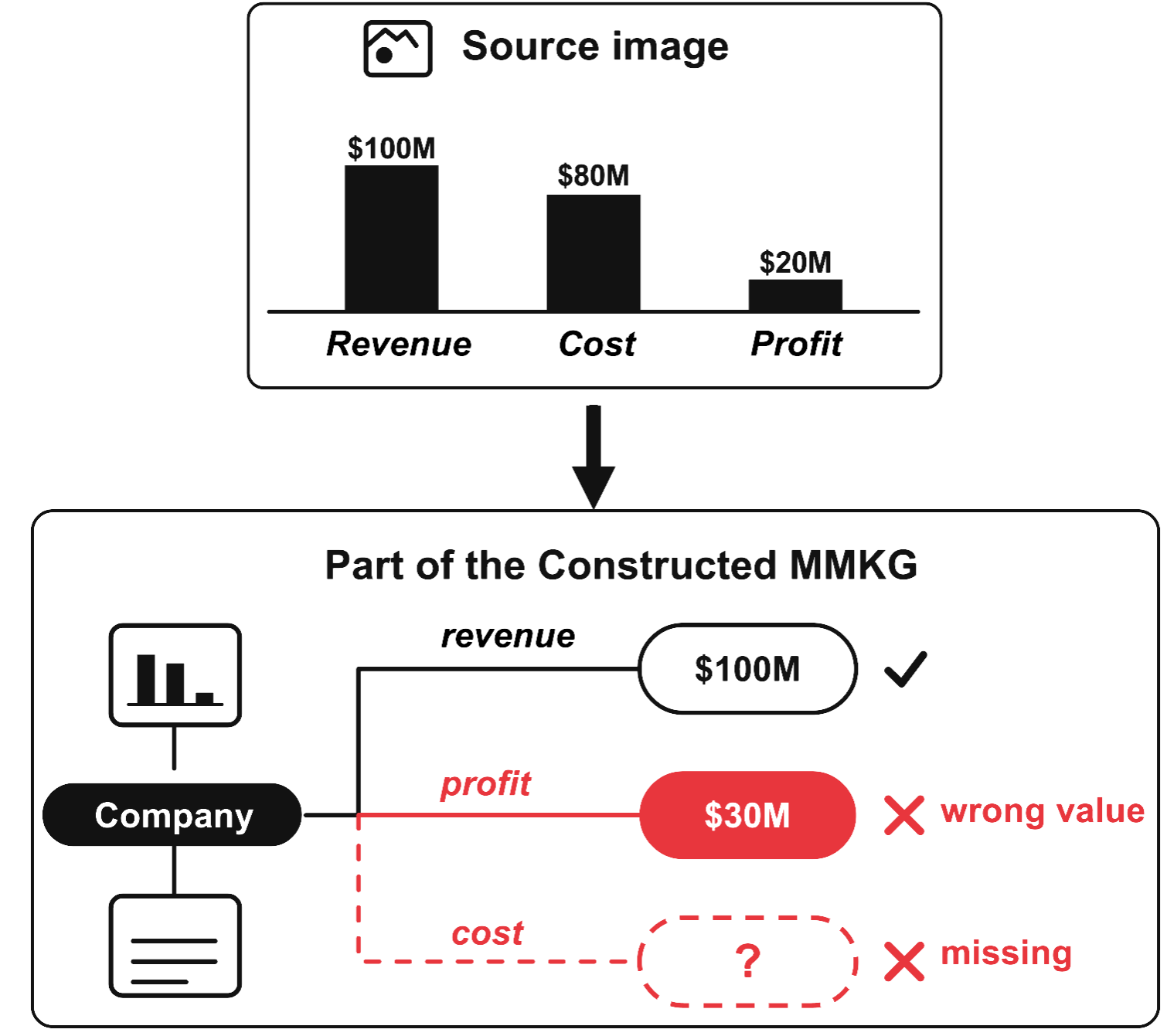}
    \caption{An example illustrating that a correct downstream answer for a certain question does not necessarily indicate that all content in the constructed MMKG is accurate or of high quality.  MMKG-R1 introduces explicit quality-aware rewards to assess the constructed MMKG and utilize them to train a multimodal constructor.}
    \label{fig:intro}
\end{figure}

Although MMKG-based Graph RAG methods have shown promising performance, most existing approaches~\citep{guo2025raganything,wan2025mmgraphrag} follow a common pipeline in which MMKGs are constructed directly by pretrained models without explicitly considering their utility for downstream RAG tasks. Since the performance of these methods heavily rely on the quality of the constructed MMKGs, a promising direction is to train a model to generate high-quality MMKGs. 

Some recent approaches~\citep{tsang2026autographr1,Lin2026EvoGraphR1} have begun exploring graph optimization through reinforcement learning (RL). AutoGraph-R1~\citep{tsang2026autographr1} trains constructors for text-derived KGs using task-aware rewards based on deducibility and passage recall. EvoGraph-R1~\citep{Lin2026EvoGraphR1} trains an agent to retrieve from and edit a multimodal knowledge hypergraph, primarily guided by final answer corretness. However, as noted in AutoGraph-R1, answer-level feedback alone does not reliably reflect graph quality. A correct downstream answer does not necessarily imply that the MMKG correctly represents the source documents. For example, as shown in Figure~\ref{fig:intro}, a model using multimodal Graph RAG may correctly answer a question about a company's profit even when the constructed MMKG records an incorrect profit value. Other correctly extracted information, such as the company's revenue, may still enable retrieval of the relevant source image, allowing the model to obtain the correct profit value directly from the image. However, such noise in the MMKG may result in reduced performance on other questions. For example, when asked "What is the revenue of the company with a profit of \$20M?", the system may fail to identify the relevant company and retrieve its source image due to the incorrect profit value in the MMKG, potentially leading to an incorrect answer.

Therefore, in this paper, we propose to study a novel problem of designing reward signals that assess both the quality of information extracted from individual knowledge sources and the quality of the MMKG assembled from the extracted information to better train the MMKG constructor through RL. This raises two key challenges: (\textbf{i}) How can we directly evaluate the quality of constructed MMKGs without relying on final answer correctness? (\textbf{ii}) How can we provide effective reward signals at both the knowledge-source and MMKG levels to guide constructor training? To address the challenges, we propose MMKG-R1, a novel graph-quality-driven RL framework that trains a multimodal constructor using complementary feedback at both MMKG and knowledge-source levels. At the MMKG level, we design \emph{Deducibility Reward} to evaluate the evidence retrieved from the assembled MMKG, encouraging the constructor to preserve relevant facts and relations from the original knowledge sources. At the knowledge-source level, we design two reward signals. The \emph{Source-Grounded Quality Reward} compares extracted information with its original knowledge sources to evaluate extraction accuracy and coverage of key information, independent of the current question. The \emph{Graph-Readiness Reward} encourages well-structured extractions that provide usable entities and facts for MMKG assembly.

Our \textbf{main contributions are}: \textbf{(i)} We present MMKG-R1, a novel quality-driven RL framework for constructing high-quality MMKGs from multimodal knowledge sources; \textbf{(ii)} We move beyond conventional final-answer-based rewards by designing multiple quality-aware reward signals that directly assess construction quality at both the knowledge-source and MMKG levels, providing more targeted feedback to guide MMKG construction; and \textbf{(iii)} We conduct comprehensive experiments on multiple datasets, showing that MMKG-R1 consistently generates high-quality MMKGs that can effectively support multimodal Graph RAG.

\section{Related Work}
\subsection{Graph RAG}
Retrieval-augmented generation (RAG)~\citep{Lewis2020RAG,chen2022murag,hu2023reveal,yu2025visrag,abootorabi2025mmragsurvey,gao2024llmrag} provides external evidence to mitigate the knowledge limitations and hallucinations of (multimodal) large language models. Graph RAG extends this paradigm by organizing knowledge into graph structures to support retrieval and generation~\citep{edge2024localglobalgraphrag,sun2024tog,guo2025lightrag,wang2026GPR}. For example, GraphRAG~\citep{edge2024localglobalgraphrag} uses hierarchical community summaries, while LightRAG~\citep{guo2025lightrag} combines graph-based indexing with dual-level retrieval. Recent methods construct multimodal knowledge graphs (MMKGs) to support multimodal Graph RAG~\citep{wan2025mmgraphrag,xu2026mkgrag,guo2025raganything,dai2026mg2rag,wu2026mmkgcontext}. For instance, MMGraphRAG~\citep{wan2025mmgraphrag} builds image-based entity-relation graphs and a text-based KG, respectively, then integrates them through cross-modal fusion. However, most existing methods rely on pretrained models without task-specific training of the MMKG construction process.

\subsection{Reinforcement Learning for Graph Construction}
Reinforcement learning (RL) methods~\citep{schulman2017ppo} like Group Relative Policy Optimization (GRPO)~\citep{shao2024grpo}, has been used to train LLMs for retrieval decisions and graph reasoning~\citep{jin2025searchr1,wu2026mmsearchr1,luo2026graphr1,xu2026veritas,zhang2025rlreasonkg}. Recent studies further apply RL to graph construction. AutoGraph-R1~\citep{tsang2026autographr1} uses GRPO to train a textual graph constructor with graph utility rewards, while EvoGraph-R1~\citep{Lin2026EvoGraphR1} learns to evolve query-conditioned multimodal hypergraphs using rewards for answer correctness, trajectory structure, and action cost. However, rewards in EvoGraph-R1 do not directly evaluate the quality of the constructed graph. MMKG-R1 instead learns to construct high-quality multimodal knowledge graphs through direct quality feedback. Our designed rewards assess both the information extracted from individual sources and the assembled MMKG, providing supervision for source-grounded extraction and graph readiness.
\section{Preliminaries}
\label{sec:preliminaries}
In this section, we first formalize how MMKG-R1 constructs an MMKG and how the constructed graph supports question answering, and then define the learning problem.

\paragraph{Multimodal graph construction.}
Let $\mathcal{X}$ denote a collection of source instances from multimodal documents, each being a text passage, an image, or a table. MMKG-R1 trains an MLLM $\pi_\theta$ as the \emph{multimodal constructor} to generate a structured extraction $y_x$ of typed entities and subject–relation–object triples from the visible content $v(x)$ of each source $x \in \mathcal{X}$:
\begin{equation}
    y_x \sim \pi_\theta(\cdot \mid v(x)).
    \label{eq:extraction}
\end{equation}
Additional source information $\mu_x$, such as provenance metadata and references to the original content, is maintained by the environment and is never shown to $\pi_\theta$. Two fixed, non-trainable modules then process the extractions: the parser $\mathcal{P}$ parses each $y_x$, and the assembler $\mathcal{B}$ uses the parsed results and $\mu_x$ to assign graph identifiers, attach provenance, link entities across sources, and assemble the graph as:
\begin{equation}
    G_\theta(\mathcal{X}) = \mathcal{B}\big(
    \bigl\{(\mathcal{P}(y_x),\mu_x)\bigr\}_{x\in\mathcal{X}}
    \big).
    \label{eq:graph}
\end{equation}
The resulting $G_\theta$ is a \emph{fact-centric} and \emph{source-grounded} MMKG: each extracted triple becomes a fact node connected to its arguments and its supporting source, and fixed entity linking connects compatible mentions across sources. Since $\mathcal{P}$ and $\mathcal{B}$ are fixed, the quality of $G_\theta$ is governed by the extractions of $\pi_\theta$.

\paragraph{Graph-based question answering.}
Given a question $q$ and a constructed MMKG $G$, a fixed retrieval and evidence-packing pipeline selects relevant evidence under predefined budgets:
\begin{equation}
    E(q,G) = (\mathcal{F}_q, \mathcal{V}_q),
    \label{eq:evidence}
\end{equation}
where $\mathcal{F}_q$ contains retrieved structured facts with source provenance, and $\mathcal{V}_q$ contains selected original visual inputs (images and image-backed tables) with their available source metadata. A downstream model answers $q$ using only $E(q,G)$, without access to the full set of sources.

\paragraph{Problem formulation.}
Each training example $\xi_q=(q,\mathcal{A}_q,\mathcal{X}_q)\in\mathcal{D}$ consists of a question $q$, a set of accepted answers $\mathcal{A}_q$, and the annotated supporting sources $\mathcal{X}_q$. Construction is \emph{question-blind}: $\pi_\theta$ processes each $x\in\mathcal{X}_q$ independently from $v(x)$ alone, without access to $q$, $\mathcal{A}_q$, or the other sources, and yields $G_\theta(\mathcal{X}_q)$ via Eqs.~\eqref{eq:extraction}--\eqref{eq:graph}, so its extractions can be reused across questions. The question and answers are used only by the training environment to assess the resulting graph. Therefore, we seek to optimize
\begin{equation}
    \max_{\theta}\;
    \mathbb{E}_{\xi_q\sim\mathcal{D}}\,
    \mathbb{E}_{\pi_\theta}
    \Big[\mathcal{Q}\big(G_\theta(\mathcal{X}_q);\,\xi_q\big)\Big],
    \label{eq:objective}
\end{equation}
where the inner expectation is over the sampled extractions $\{y_x\}_{x\in\mathcal{X}_q}$, and $\mathcal{Q}$ scores the quality of the graph itself rather than the correctness of a final answer. Section~\ref{sec:method} specifies $\mathcal{Q}$ and how Equation.~\eqref{eq:objective} is optimized.

\section{Method}
\label{sec:method}

\begin{figure*}[t]
    \centering
    \includegraphics[width=1\linewidth]{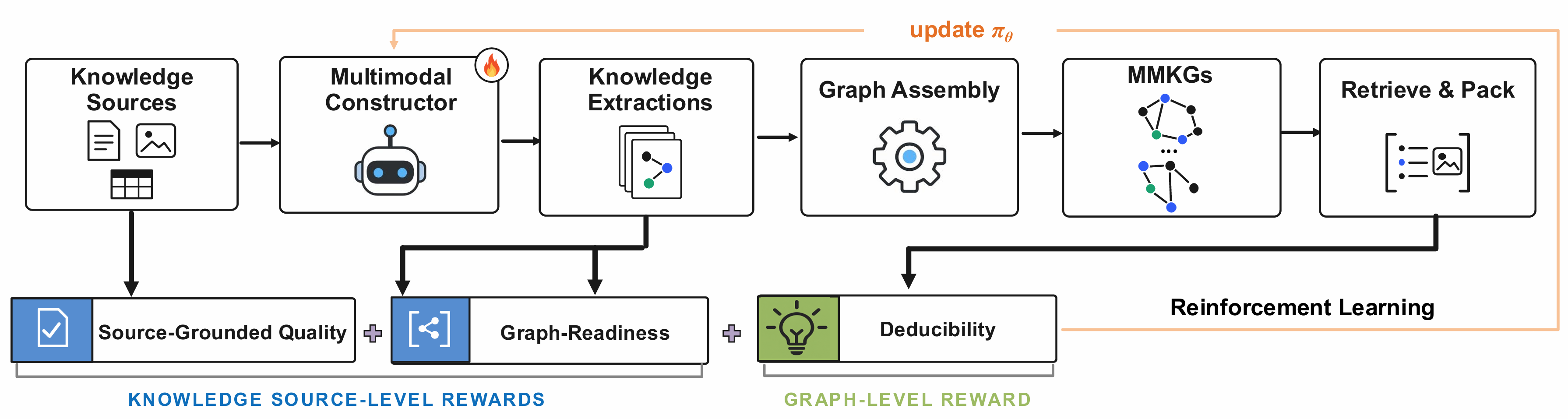}
    \caption{Overview of MMKG-R1. The constructor independently extracts structured knowledge from each knowledge source. The extracted knowledge is then used to assemble MMKGs. Three designed rewards guide constructor optimization: the graph-level Deducibility Reward and two source-level rewards, the Source-Grounded Quality Reward and the Graph-Readiness Reward.}
    \label{fig:method}
\end{figure*}

Our proposed MMKG-R1 trains a multimodal constructor to facilitate the contruction of MMKGs that effectively support downstream question answering. As shown in Figure~\ref{fig:method}, we introduce three complementary reward signals that jointly evaluate evidence sufficiency for downstream question answering, fidelity to the original knowledge source, and structural usability: (i) \textbf{Deducibility Reward}, which evaluates whether the retrieved evidence from a constructed MMKG is sufficient to support a correct answer; (ii) \textbf{Source-Grounded Quality Reward}, which assesses whether the knowledge extracted by the constructor faithfully preserves the central content of its original knowledge source; and (iii) \textbf{Graph-Readiness Reward}, which encourages structurally valid and usable outputs for graph construction. Evidence sufficiency depends on the question and on how extractions from different sources combine, so the Deducibility Reward is computed at the candidate-graph level. Fidelity and usability are properties of each extraction regardless of the question, so the Source-Grounded Quality and Graph-Readiness Rewards are computed at the source-extraction level.

\subsection{Source-Level Construction and Graph Assembly}
\label{sec:graph_construction}

For each training question $q$, we consider its annotated supporting knowledge sources $\mathcal{X}_q$, as defined in
Section~\ref{sec:preliminaries}. The constructor $\pi_\theta$ processes each knowledge source $x_m^{(q)}$ independently, using only its content $v(x_m^{(q)})$. Let $\pi_{\theta_{\mathrm{old}}}$ denote the rollout policy
used to collect the current training batch.
For each knowledge source $x_m^{(q)}$, we independently
sample four responses $y_{m,i}^{(q)} \sim \pi_{\theta_{\mathrm{old}}}(\cdot \mid v(x_m^{(q)}))$, where $i \in \{1,2,3,4\}$.

Constructor outputs with the same sampling index $i$ are then parsed and combined into a candidate multimodal knowledge graph $G_i^{(q)}$, while preserving source metadata. This produces four candidate MMKGs per question for group-relative comparison, without enumerating all possible combinations of outputs across knowledge sources. For each candidate graph $G_i^{(q)}$, we use the fixed retrieval and evidence-packing pipeline introduced in Section~\ref{sec:preliminaries} to obtain $E_i^{(q)} = E(q, G_i^{(q)})$, which includes structured facts retrieved from the graph and original visual evidence when available. We use a shared, fixed environment to perform entity linking, graph assembly, retrieval, and budgeted evidence packing for all candidates. Throughout training, only the constructor is updated, while all other components remain fixed.

\subsection{Graph-Level Deducibility Reward}
\label{sec:deducibility}
We first present the design of the Deducibility Reward, which is inspired by the Knowledge-Carrying Reward proposed in AutoGraph-R1~\citep{tsang2026autographr1}. For each candidate MMKG $G_i^{(q)}$, we use a frozen MLLM as the Judge to evaluate whether the retrieved evidence $E_i^{(q)}$ from $G_i^{(q)}$ provides sufficient support for at least one accepted complete answer in $\mathcal{A}_q$. Training graphs are constructed from the annotated supporting sources $\mathcal{X}_q$, which contain the evidence needed to answer $q$.

To be specific, the judge model is prompted to evaluate two types of evidence gaps in sequence: (i) \emph{Whole-bundle gap}: Indicates that both the retrieved
structured facts and the original visual information in knowledge sources are insufficient to support any complete accepted answer; and (ii) \emph{Structured-path gap}: This gap is evaluated only when the whole bundle is sufficient. It indicates that omissions or contradictions in answer-critical information prevent the structured facts from supporting a complete, correct, and unambiguous reasoning path. The two checks distinguish the usefulness of the retrieved evidence from the completeness and correctness of the structured reasoning path.

Each reported gap describes concrete missing or conflicting evidence. A reported whole-bundle gap results in a score of $0$. When only a structured-path gap exists, the evidence supports an answer, but the structured reasoning path is incomplete or incorrect. We assign a score of $0.5$ to recognize the usefulness of the evidence while encouraging improvements to the structured facts. A candidate with neither gap receives a score of $1$. These scores define three levels of evidence utility, formalized as follows:
\begin{equation}
d_i^{(q)} =
\begin{cases}
0, & \text{whole-bundle gap},\\
0.5, & \text{structured-path gap only},\\
1, & \text{neither gap}.
\end{cases}
\label{eq:deducibility}
\end{equation}

\subsection{Knowledge Source-Level Feedback}
\label{sec:source-feedback}
While the Deducibility Reward described above evaluates whether retrieved graph evidence is sufficient to support downstream question answering, it does not fully capture the quality of individual knowledge source-level extractions. To address this limitation, we introduce two complementary knowledge source-level rewards: the \emph{Source-Grounded Quality Reward} and the \emph{Graph-Readiness Reward}.

\paragraph{Source-Grounded Quality Reward.}
\label{sec:source-quality}
The proposed Deducibility Reward evaluates evidence for the current question, but source extractions are intended for reuse across queries. An extraction may retain enough information to retrieve a relevant image or answer the current question while omitting other central content or introducing errors unrelated to that question. These deficiencies may leave the Deducibility Reward unchanged, yet limit the graph's reliability and usefulness for other queries. We therefore introduce the Source-Grounded Quality Reward to encourage the constructor to accurately extract the knowledge source's central knowledge. A frozen judge model is prompted to evaluate each selected extraction against its original knowledge source, without access to the question or accepted answers.

The judge model evaluates two complementary criteria:
(\textbf{i}) \emph{reliability}, requiring the extraction
to contain at least one meaningful, source-supported
assertion and no material factual or binding errors; and (\textbf{ii}) \emph{central coverage}, requiring the extraction to capture the source's key explicit information, including entities, relations, values, and necessary qualifiers, without requiring exhaustive coverage of minor details.

An extraction receives a binary quality reward
$u_{m,i}^{(q)}=1$ only if both criteria are satisfied;
otherwise, $u_{m,i}^{(q)}=0$. Thus, an accurate extraction that omits central information, or a comprehensive extraction containing material errors, receives a reward of $0$. Source-Grounded Quality Reward is considered only when
all four candidate graphs have valid judge evaluations and
receive the same Deducibility Reward of either $0.5$ or $1$.
In these groups, the retrieved evidence is sufficient to support a complete answer, so the graph-level reward cannot distinguish among candidates. Source-level evaluation then provides additional feedback by assessing extraction accuracy and coverage of key information that the question-specific reward may overlook. This encourages the constructor to better preserve information from the original sources, even when candidates are equally answerable. For each group, we sample at most one eligible source to assess extraction quality, providing additional supervision while limiting evaluation costs.

\paragraph{Graph-Readiness Reward.}
\label{sec:graph-readiness}

To encourage the constructor to generate structurally
valid and usable extractions, we introduce the
Graph-Readiness Reward. It considers both the format
validity of the original response, similar to format
rewards used in prior
work~\citep{jin2025searchr1extension},
and the usability of the extracted content for MMKG
construction.

For each response $y_{m,i}^{(q)}$, we define a binary
Graph-Readiness reward $r_{m,i}^{(q)} \in \{0,1\}$.
We set $r_{m,i}^{(q)}=1$ only if the original output
satisfies the required JSON schema without repair,
the response terminates normally and passes the
generation-protocol checks, and it produces at least
one usable source-bound entity mention or fact.
Otherwise, $r_{m,i}^{(q)}=0$.
Malformed responses receive a reward of $0$ even if
the parser can recover their content, as do
structurally valid but empty responses.

\begin{table*}[t]
\centering
\small
\renewcommand{\arraystretch}{1.12}
\begin{tabularx}{\textwidth}{Xccc}
\toprule
\textbf{Method} & \textbf{MMQA} & \textbf{MMDocRAG} & \textbf{MMLongBench-Doc} \\
\midrule
Direct Inference & 23.96 & 50.45 & 24.73 \\
Naive RAG & 29.79 & 59.46 & 25.38 \\
\midrule
Base GraphRAG (Dense) & 33.75 & 67.12 & 40.65 \\
F1-based RL (Dense) & 35.83 & 71.17 & 40.43 \\
\rowcolor[RGB]{237,244,248}
\textbf{MMKG-R1} (Dense)  & \textbf{38.13} & \textbf{72.07} & \textbf{41.29} \\
\midrule
Base GraphRAG (Subgraph) & 29.58 & 63.96 & 40.65 \\
F1-based RL (Subgraph) & 29.79 & 64.41 & 41.08 \\
\rowcolor[RGB]{237,244,248}
\textbf{MMKG-R1} (Subgraph)  & \textbf{33.96} & \textbf{65.32} & \textbf{41.51} \\
\midrule
Base GraphRAG (Hybrid) & 40.83 & 66.67 & 41.94 \\
F1-based RL (Hybrid) & 38.54 & 68.02 & 42.15 \\
\rowcolor[RGB]{237,244,248}
\textbf{MMKG-R1} (Hybrid)  & \textbf{42.50} & \textbf{68.92} & \textbf{42.37} \\
\bottomrule
\end{tabularx}
\caption{Answer accuracy (\%) on the three evaluation subsets. Qwen3-VL-8B-Instruct is used as the answer model. Direct Inference and Naive RAG do not construct knowledge graphs. For MMLongBench-Doc, trained constructors are transferred from MMDocRAG without additional training. Bold indicates the best result within each dataset--graph-retrieval block.}
\label{tab:main-results}
\end{table*}

\subsection{GRPO-Based Optimization}
\label{sec:optimization}
We normalize Deducibility rewards $d_i^{(q)}$ across
valid candidate graphs for each question to compute
the advantages $A_{m,i}^{(q),D}$ for their corresponding
source responses. The source-level advantages
$A_{m,i}^{(q),Q}$ and $A_{m,i}^{(q),R}$ are computed
from separately mean-centered rewards $u_{m,i}^{(q)}$
and $r_{m,i}^{(q)}$ across responses generated for
the same source. After applying the corresponding activation and source-credit masks, the combined advantage for each source response \(y_{m,i}^{(q)}\) is:
\begin{equation}
A_{m,i}^{(q)}
= A_{m,i}^{(q),D}
+ \lambda_Q A_{m,i}^{(q),Q}
+ \lambda_R A_{m,i}^{(q),R},
\label{eq:combined}
\end{equation}
where $\lambda_Q$ and $\lambda_R$ weight the two source-level advantages. Finally, we optimize the constructor with GRPO by maximizing a dual-clipped surrogate objective~\citep{ye2020mastering}.
\section{Experiments}

In this section, we conduct comprehensive experiments to answer these research questions: (i) \textbf{RQ1}: Can our proposed MMKG-R1 construct high-quality multimodal knowledge graphs (MMKGs) that improve the performance of multimodal Graph RAG? (ii) \textbf{RQ2}: Are our proposed reward scores more effective than the F1 score in guiding the construction of high-quality MMKGs? and (iii) (\textbf{RQ3}): How do different reward designs affect the performance of MMKG-R1?

\subsection{Experimental Setup}
\label{sec:experimental-setup}
\subsubsection{Datasets} 
We train two multimodal knowledge constructors separately on the training sets of MMDocRAG~\citep{dong2025MMDocRAG} and MultiModalQA (MMQA)~\citep{talmor2021multimodalqa}. We then evaluate them on subsets of their respective test sets. We use subsets for two reasons: (\textbf{i}) Some test questions share gold passages with training questions, which may lead to data leakage and affect evaluation reliability; (\textbf{ii}) Some test questions can be answered using textual information alone, without requiring visual evidence, making them less suitable for evaluating multimodal knowledge construction. To assess the transferability of MMKG-R1, we further evaluate its performance on a subset of MMLongBench-Doc~\citep{ma2024MMLongBenchDoc}. The details of all used datasets are described in Appendix~\ref{app:dataset}.

\subsubsection{Evaluation Protocol} 
We use a judge model to evaluate answer accuracy. To reduce potential evaluation bias, we use Qwen3.5-27B~\citep{qwen35blog} as an independent judge, distinct from the model used for calculating rewards during training. For each dataset, all methods are evaluated using the same prompt and evaluation criteria.

\begin{table}[t]
\centering
\small
\setlength{\tabcolsep}{3pt}
\renewcommand{\arraystretch}{1.08}
\begin{tabular*}{\columnwidth}{@{\extracolsep{\fill}}llrrr@{}}
\toprule
\textbf{Answer Model} & \textbf{Method} & \textbf{Dense} & \textbf{Subgraph} & \textbf{Hybrid} \\
\midrule
\multirow{3}{*}{32B-AWQ}
 & Base & 39.79 & 35.42 & 43.96 \\
 & F1-based RL & 40.00 & 36.46 & 46.46 \\
 & MMKG-R1 & \textbf{40.63} & \textbf{38.54} & \textbf{48.96} \\
\bottomrule
\end{tabular*}
\caption{Performance of MMKG-R1 on MMQA with Qwen3-VL-32B-Instruct-AWQ as the answer model. Base denote Base Graph RAG.}
\label{tab:answer-model-transfer}
\end{table}

\subsubsection{Baseline Methods}
We compare MMKG-R1 with four baseline methods: (i) \textbf{Direct Inference with Materials.} We retain the supporting materials and add distractor within the model's context budget. The frozen answer model directly answers from this expanded multimodal context. (ii) \textbf{Naive RAG.} We index raw text and table chunks together with visual materials in a corpus-wide pool using a multimodal embedding model. For each question, the top 20 units are retrieved by embedding similarity and supplied to the frozen answer model. (iii) \textbf{Base GraphRAG.} A constructor without reinforcement learning extracts knowledge to build an MMKG for downstream retrieval and question answering. (iv) \textbf{F1-based RL.} This baseline adopts a correctness-oriented RL objective used in representative related work~\citep{Lin2026EvoGraphR1}, computing token-level F1 between the responses generated by a frozen answer model and the reference answers. It retains the Graph-Readiness Reward but excludes the Source-Grounded Quality Reward.

\subsubsection{Implementation Details}
\paragraph{Computational Devices.} All training and evaluation experiments are conducted on machines equipped with four NVIDIA RTX A6000 GPUs, each with 48 GB of memory. 

\paragraph{Backbone Models.} We use Qwen3-VL-4B-Instruct~\citep{bai2025qwen3vl} as the multimodal constructor. Both the Deducibility Reward and Source-Grounded Quality Reward use frozen Qwen3-VL-32B-Instruct-AWQ. The Graph-Readiness Reward is computed using deterministic rules. The downstream answer model is Qwen3-VL-8B-Instruct~\citep{bai2025qwen3vl} for all three datasets.
All methods use the same frozen answer model and answer-evaluation protocol within each dataset. 

\begin{figure}[t]
\centering
\includegraphics[width=1\columnwidth]{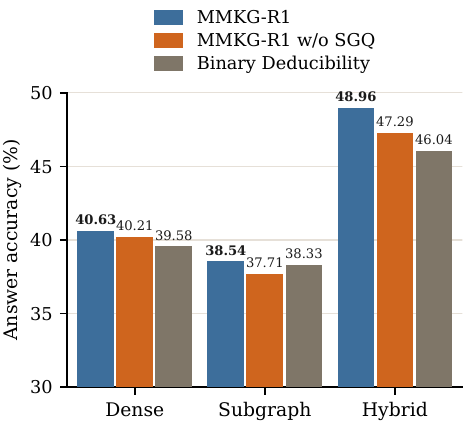}
\vspace{-10pt}
\caption{Reward ablations on MMQA subset. Metric is answer accuracy (\%).}
\vspace{-10pt}
\label{fig:reward-ablations}
\end{figure}

\begin{figure*}[t]
\centering
\includegraphics[width=1\textwidth]{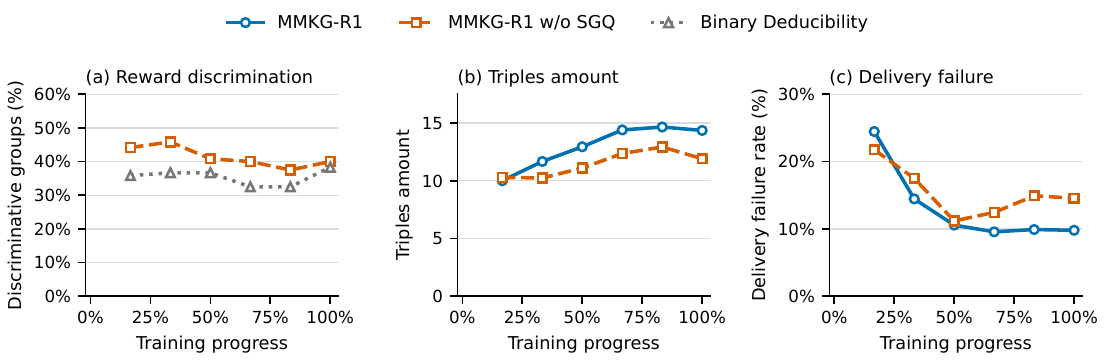}
\vspace{-10pt}
\caption{Training dynamics on MMQA. (a) Proportion of discriminative question groups during training with our Deducibility Reward and its binary variant. (b,c) Mean number of triples per source output and output delivery failure rates for MMKG-R1 and MMKG-R1 w/o SGQ.}
\label{fig:reward-training-dynamics}
\end{figure*}

\paragraph{Retrieval Settings}
During training, we use a fixed retrieval and evidence-packing procedure with entity, fact, and relation entry points to compute the Deducibility Reward. At evaluation time, we assess the same trained constructor under three different retrieval methods: \emph{Dense Retrieval}, \emph{Subgraph Retrieval}, and \emph{Hybrid Retrieval}.

More implementations are described in~\ref{app:implementdetails}.
\subsection{Main Results}
Table~\ref{tab:main-results} reports answer accuracy across the three evaluation subsets. Direct Inference and Naive RAG generally achieve lower accuracy than graph-based retrieval methods. This highlights the benefits of structured multimodal knowledge in graphs. F1-based RL training improves downstream accuracy over Base GraphRAG in most settings, showing that final-answer correctness can provide useful feedback for MMKG construction. However, these improvements are not consistent across all settings. For example, on the MMQA subset with hybrid retrieval, accuracy drops from 40.83\% for the untrained Base GraphRAG to 38.54\% after F1-based RL training, suggesting that answer-based rewards may not always provide reliable guidance for MMKG construction. Our proposed MMKG-R1 consistently outperforms both Base GraphRAG and F1-based RL, For example, on the MMQA subset with dense retrieval, MMKG-R1 achieves an accuracy of 38.13\%, outperforming both Base GraphRAG (33.75\%) and F1-based RL (35.83\%). This demonstrates that our quality-aware rewards provide effective guidance for MMKG construction

\subsection{Transferability of MMKG-R1}
\paragraph{Different Answer Models.}
On the MMQA subset, we additionally evaluate Qwen3-VL-32B-Instruct-AWQ as the answer model, alongside Qwen3-VL-8B-Instruct. The results are shown in Table~\ref{tab:answer-model-transfer}. MMKG-R1 consistently constructs MMKGs that achieve higher answer accuracy across all three retrieval strategies, even when evaluated with a different answer model.

\paragraph{Cross-Dataset Transferability.} To evaluate the transferability of MMKG-R1, We further evaluate the MMDocRAG-trained constructors on MMLongBench-Doc without further training. As shown in Table~\ref{tab:main-results}, MMKG-R1 consistently outperforms both Base GraphRAG and F1-based RL across all three retrieval strategies. These results suggest that our quality-aware rewards help the constructor learn knowledge construction capabilities that transfer to unseen datasets.

\subsection{Ablation Studies}
\label{sec:ablation-studies}

To examine the contributions of our reward design, we evaluate two variants on the MMQA subset: (i) \emph{MMKG-R1 w/o Source-Grounded Quality (SGQ) Reward.} This variant removes the SGQ Reward while retaining the Deducibility and Graph-Readiness Rewards. (ii) \emph{Binary Deducibility.} This variant removes the SGQ Reward and binarizes the Deducibility Reward by mapping both $0.5$ and $1$ to $1$, while keeping $0$ unchanged. It no longer distinguishes between incomplete and complete structured reasoning paths when the evidence supports the answer. We use Qwen3-VL-32B-Instruct-AWQ for answer generation and evaluation in this ablation study. As shown in Figure~\ref{fig:reward-ablations}, full MMKG-R1 consistently outperforms both variants. Removing the SGQ Reward reduces accuracy, while binarizing the Deducibility Reward generally leads to further degradation. These results suggest that MMKG-level deducibility feedback alone does not fully substitute for source-grounded quality supervision. We further examine how our Deducibility Reward compares with its binary variant during training, and how adding SGQ affects the constructor's outputs.

\paragraph{Deducibility Reward vs. Its Binary Variant.}
We first compare how our Deducibility Reward and its binary variant distinguish candidate graphs for the same question. A group is discriminative if at least three valid candidates remain and their task rewards are not all identical. Figure~\ref{fig:reward-training-dynamics}(a) reports the proportion of such groups among all sampled groups, with SGQ disabled in both variants. Our reward yields a higher proportion across all reported windows. Unlike the binary variant, it distinguishes incomplete from complete structured reasoning paths even when both support the answer, providing finer feedback for relative optimization.

\paragraph{Effect of SGQ.}
We next examine how adding SGQ affects the constructor's outputs using two metrics: (1) \emph{Triples amount}, defined as the average number of triples obtained from each source output after parsing and deduplication; and (2) \emph{Output delivery failure rate}, defined as the proportion of source outputs that fail the structural, normal-termination, or nonempty-output requirements. Figure~\ref{fig:reward-training-dynamics}(b,c) compares these metrics throughout training between full MMKG-R1 and MMKG-R1 w/o SGQ. As training progresses, the full model retains more triples while exhibiting fewer output delivery failures. These observations complement the downstream accuracy gains, suggesting that SGQ helps preserve structured information and improve output reliability.

Together, these analyses support the complementary roles of the two rewards: our Deducibility Reward provides differentiated task feedback among candidates for the same question, while SGQ supplies additional quality supervision at the source level. Both contribute to training the constructor.

\section{Conclusion}
We introduce MMKG-R1, a novel reinforcement learning framework that trains a multimodal constructor to help build useful multimodal knowledge graphs (MMKGs) for downstream question answering. MMKG-R1 combines the graph-level Deducibility Reward with two knowledge-source-level rewards, the Source-Grounded Quality Reward and the Graph-Readiness Reward, to provide complementary feedback on evidence sufficiency, source fidelity, and usability. Results across multiple datasets show that the multimodal constructor trained with MMKG-R1 can effectively extract knowledge to build useful MMKGs, while demonstrating strong generalization to unseen datasets.

\section*{Limitations}
The Deducibility and Source-Grounded Quality rewards rely on external multimodal large language models as judges, making their reliability dependent on the quality of the judge model's assessments.

\section*{Ethical considerations}
This work aims to train a multimodal large language model to extract high-quality knowledge from multimodal sources for constructing multimodal knowledge graphs. All experiments are conducted based on publicly available datasets. We do not anticipate significant ethical or societal risks arising from this research.


\bibliography{custom}

\appendix

\section{Dataset Details}
\label{app:dataset}
\paragraph{MMDocRAG.}
MMDocRAG~\citep{dong2025MMDocRAG} is designed to evaluate multimodal evidence retrieval and question answering over multimodal documents. MMDocRAG is licensed under the Apache License 2.0.

\paragraph{MultiModalQA.}
MultiModalQA (MMQA)~\citep{talmor2021multimodalqa} is built from Wikipedia tables linked to text passages and images, and is designed to evaluate complex question answering involving cross-modal and multi-hop reasoning. In our experiments, text and images are preserved in their original forms, while structured tables are serialized into text. MMQA is publicly available, but no explicit license is specified.

\paragraph{MMLongBench-Doc.}
MMLongBench-Doc~\citep{ma2024MMLongBenchDoc} evaluates long-document understanding involving information like text and images. We select answerable questions with valid supporting pages and evidence types that include figures or charts. This subset is used only for cross-dataset transfer evaluation with the constructor trained on MMDocRAG. MMLongBench-Doc is licensed under the CC BY-NC 4.0.

\section{More Implementation Details}
\label{app:implementdetails}
For constructor rollouts during the training phase, we set the sampling temperature to 1.0 and the maximum output length to 4,096 tokens. We build our training pipeline using verl~\citep{sheng2025verl} and use vLLM~\citep{Kwon2023vllm} for answer model inference.

\section{AI Usage Disclosure}
The authors used generative AI tools to assist with writing and coding. All AI-assisted content and code were reviewed and verified by the authors
\end{document}